\documentclass{article}
\newcommand{\remove}[1]{}

\newcommand{\network}{g}

\newcommand{\sloss}{\bar{\ell}}

\newcommand{\adv}[1]{\tilde{#1}}
\newcommand{\perturb}[1]{\delta_{#1}}
\newcommand{\adveps}{\epsilon}

\newcommand{\R}{\mathbb{R}}
\newcommand{\N}{\mathbb{N}}

\newcommand{\x}{\mathbf{x}}
\newcommand{\X}{\mathbf{X}}
\newcommand{\Un}{\mathbf{U}}
\newcommand{\M}{\mathbf{M}}
\newcommand{\Z}{\mathbf{Z}}
\newcommand{\z}{\mathbf{z}}

\newcommand{\argmin}{\arg \min}

\newcommand{\figref}[1]{Figure~\ref{#1}}
\newcommand{\secref}[1]{Section~\ref{#1}}

\newcommand{\len}{len}

\usepackage[preprint,nonatbib]{nips_2018}

\usepackage[utf8]{inputenc} % allow utf-8 input
\usepackage[T1]{fontenc}    % use 8-bit T1 fonts
\usepackage{hyperref}       % hyperlinks
\usepackage{url}            % simple URL typesetting
\usepackage{booktabs}       % professional-quality tables
\usepackage{amsfonts}       % blackboard math symbols
\usepackage{nicefrac}       % compact symbols for 1/2, etc.
\usepackage{microtype}      % microtypography

\usepackage{hyperref,graphicx}
\usepackage{amsmath,amsthm,amssymb,bbm,bm}
\usepackage{url}
\usepackage{subcaption}
\usepackage[ruled,norelsize]{algorithm2e}
\usepackage{color}
\usepackage{soul}
\usepackage[square,sort,comma,numbers]{natbib}

\title{Deceiving End-to-End Deep Learning Malware Detectors using Adversarial Examples}
\author{
  Felix Kreuk, Assi Barak, Shir Aviv, Moran Baruch, Benny Pinkas, Joseph Keshet\\
  Dept. of Computer Science,
  Bar-Ilan University, Israel \\
  \texttt{felixkreuk@gmail.com} \\
}

\begin{document}

\maketitle

\begin{abstract}
In recent years, deep learning has shown performance breakthroughs in many prominent problems. However, this comes with a major concern: deep networks have been found to be vulnerable to adversarial examples. Adversarial examples are modified inputs designed to cause the model to err. In the domains of images and speech, the modifications are untraceable by humans, but affect the model's output. In binary files, small modifications to the byte-code often lead to significant changes in its functionality/validity, therefore generating adversarial examples is not straightforward. We introduce a novel approach for generating adversarial examples for discrete input sets, such as binaries. We modify malicious binaries by injecting a small sequence of bytes to the binary file. The modified files are detected as benign, while preserving their malicious functionality. We applied this approach to an end-to-end convolutional neural malware detector and present a high evasion rate. Moreover, we showed that our payload is transferable within different positions of the same file and across different files and file families.
\end{abstract}

% !TEX root = main.tex

\section{Introduction}
\label{sec:Intro2}

End-to-end (E2E) deep-learning has recently become ubiquitous in the field of AI. In E2E deep-learning, manual feature engineering is replaced with a unified model that maps raw input to output.
This approach has been successfully applied to many prominent problems \cite{alexnet2012,he2016deep,socher2012convolutional,amodei2016deep,bahdanau2014neural}, leading to state-of-the-art models.
Naturally, it is becoming a technology of interest in malware detection \cite{shin2015Functions,chua2017FunctionTypes,Raff2017}, where manual feature extraction is gradually being augmented/replaced by E2E models that take raw inputs.

Despite its success, deep networks have shown to be vulnerable to \emph{adversarial examples} (AEs). AEs are artificial inputs that are generated by modifying legitimate inputs so as to fool classification models. 
In the fields of image and speech, modified inputs are considered adversarial when they are indistinguishable by humans from the legitimate inputs, and yet they fool the model \cite{goodfellow2014explaining, moosavi2016deepfool, kurakin2016adversarial}. Such attacks were demonstrated in \cite{carlini2017AE, xiao2018spatially, kreuk2018fooling}.
%Fooling E2E models has been successfully done in image and speech tasks \cite{carlini2017AE, xiao2018spatially, kreuk2018fooling}.
%In these fields, modified inputs are considered adversarial when they are indistinguishable by humans from the legitimate inputs, and yet they fool the model \cite{goodfellow2014explaining, moosavi2016deepfool, kurakin2016adversarial}.
Conversely, discrete sequences are inherently different than speech and images, as changing one element in the sequence may completely alter its meaning. For example, one word change in a sentence may hinder its grammatical validity or meaning. Similarly, in a binary file, where the input is a discrete sequence of bytes, changing one byte may result in invalid byte-code or different runtime functionality.

In malware detection, an AE is a binary file that is generated by modifying an existing malicious binary. While the original file is correctly classified as \emph{malicious}, its modified version is misclassified as \emph{benign}. Recently, a series of works \cite{BGUrnn, GrossePMBM17, hu2017black} have shown that AEs cause catastrophic failures in malware detection systems. 
These works assumed a malware detection model trained on a set of handcrafted features such as file headers and API calls, whereas our work is focused on \emph{raw} binaries.
In parallel to our work, \cite{kolosnjaji2018adversarial} suggested an AE generation method for binaries by injecting a sequence of bytes. Our work differs in two main aspects: 
(i) the injection procedure in \cite{kolosnjaji2018adversarial} is based on end-of-file injections, while our method utilises multiple injection locations;
(ii) the evasion rate of \cite{kolosnjaji2018adversarial} is linearly correlated with the length of the injected sequence, while the evasion rate of our method is invariant to the length of the injected sequence.

%As malware detection techniques are becoming more data-driven, deep learning based components are gradually being used both in academia and industry \cite{Avast, Malwarebytes, Microsoft_ML_malware}. Consequently, malware creators will seek to take advantage of the vulnerability of E2E models to AEs. \ES{Seems like a broken paragraph}

%Attacks by AEs accentuate the fragility of deep learning~\cite{papernot2017practical}, this has triggered an active line of research concerned with understanding the AEs \cite{Fawzi2017robustness}, and making neural nets more robust \cite{papernot2016distillation,cisse2017parseval}.
%Numerous attempts were made at identifying and preventing such attacks \cite{yuan2017adversarial}, resulting in a game of cat-and-mouse between attackers and defenders. Defense methods were proposed, only to be later defeated by novel attacks \cite{he2017adversarial,carlini2017adversarial,yuan2017adversarial}. Consequently, training models to be robust to AEs proved to be hard and remains a topic of active research.

In this work, we analyze malware detectors based on Convolutional Neural Networks (CNNs), such as the one described in~\cite{Raff2017}. Such detectors are not based on handcrafted features, but on the entire \emph{raw binary} sequence. Results suggest that E2E deep learning based models are susceptible to AE attacks.

\textbf{Contributions.}
We re-evaluate the reliability of E2E deep neural malware detectors beyond traditional metrics, and expose their vulnerability to AEs.
We propose a novel technique for generating AEs for discrete sequences by injecting local perturbations.
As raw binaries cannot be directly perturbed, local perturbations are generated in an embedding space and reconstructed back to the discrete input space.
We demonstrate that the same payload can be injected into different locations, and can be effective when applied to different malware files and families.
% !TEX root = main.tex
\vspace{-0.1cm}

\section{Problem Setting}
\label{sec:problem_setting}

In this section, we formulate the task of malware detection and set the notations for the rest of the paper. A binary input file is composed of a sequence of bytes. We denote the set of all bytes as $\X \subseteq [0,N-1]$, where $N=256$. A binary file is a sequence of $L$ bytes $\x = (x_1,x_2,...x_L)$, where $x_i \in \X$ for all $1 \leq i \leq L$. Note that the length $L$ varies for different files, thus $L$ is not fixed. We denote by $\X^*$ the set of all finite-length sequences over $\X$, hence $\x\in\X^*$.
The goal of a malware detection system is to classify an input binary file as malicious or benign. 
%This system is, therefore, a function that gets as input a binary file $\x\in\X^*$, and outputs the probability that the binary file $\x$ is malicious. 
Formally, we denote by  $f_\theta: \X^* \rightarrow [0,1]$ the malware detection function implemented as a neural network with a set of parameters $\theta$. If the output of $f_\theta(\x)$ is greater than $0.5$ then the prediction is considered as 1 or \emph{malicious}, otherwise the prediction is considered as 0 or \emph{benign}. 

Given a malicious file, which is correctly classified as malicious, our goal is to modify it, while retaining its runtime functionality, so that it will be classified as benign. Formally, denote by $\tilde{\x}$ the modified version of the malicious file $\x$. Note that if $\x$ is classified as malicious, $f_\theta(\x) > 0.5$, then we would like to design $\tilde{\x}$ such that $f_{\theta}(\tilde{\x}) < 0.5$, and the prediction is benign.
In this work, we focus on generating AEs for E2E models, specifically, \emph{MalConv} architecture~\cite{Raff2017}, which is a CNN based malware detector. The model is composed of embedding layer denoted $\smash{\M \in \R^{N \times D}}$, mapping from $\X$ to $\Z \subseteq \R^D$. This is followed by a Gated-Linear-Unit \cite{dauphin2016language}. The output is then passed to a \emph{temporal-max-pooling} layer, which is a special case of \emph{max-pooling}. For more details see~\cite{Raff2017}.

\section{Generating Adversarial Examples}
\label{sec:adversarial}
Generating AEs is usually done by adding small perturbations to the original input in the direction of the gradient. While such methods work for continuous input sets, they fail in the case of discrete input sets, as bytes are arbitrarily represented as scalars in $[0,N-1]$. Hence, we generate AEs in a continuous embedding space $\Z$ and reconstruct them to $\X$.

\paragraph{Generation}
Given an input file that is represented in the embedded domain $\z\in\Z^*$, an AE is a perturbed version $\adv{\z} = \z + \perturb{},$
where $\perturb{}\in\Z^*$ is an additive perturbation that is generated to preserve functionality of the corresponding $\x$, yet cause $f_\theta$ to predict an incorrect label. 
We assume that $f_\theta$ was trained and the parameters $\theta$ were fixed. An adversarial example is generated by solving the following problem: $\smash{\tilde{\z} = \argmin_{\tilde{\z}:\| \tilde{\z}-\z \|_p \leq \adveps}\sloss\big(\tilde{\z},\tilde{y};\theta\big),}$
where $\tilde{y}$ is the desired target label, $\adveps$ represents the strength of the adversary, and $p$ is the norm value. 
%In~words, we would like to minimize the loss between the prediction of $f_{\theta}$ on the AE and the target label, under the constraint that the AE is similar to the original example in $p$-norm.
Assuming the loss function $\sloss$ is differentiable, 
%the authors of \cite{shaham2015understanding} proposed to take the first order Taylor expansion of $\x\mapsto\sloss(\tilde{\z},\tilde{y};\theta)$ to compute $\perturb{}$.
when using the max-norm, $\smash{p=\infty}$, the solution is $\smash{\tilde{\z} = \z - \adveps \cdot \text{sign}\big(\nabla_{\tilde{\z}} \sloss(\tilde{\z},\tilde{y};\theta)\big),}$ which corresponds to the Fast Gradient Sign Method proposed (FGSM) in \cite{goodfellow2015explaining}. When choosing $p=2$ we get $\tilde{\z} = \z - \adveps \cdot \nabla_{\tilde{\z}} \sloss(\tilde{\z},\tilde{y};\theta)$.
After $\tilde{\z}$ was crafted, we need to reconstruct the new binary file $\tilde{\x}$. The most straightforward approach is to map each $\z_i$ to its closest neighbor in the embedding matrix $\M$.
%, which is a lookup table that assigns a vector in $\Z$ to each input in $\X$. 
%While this approach works most of the time, we can further enhance it to have higher evasion rates. 

%%%%%%%%%%%%%%%%%%%%%%%%%%%%%%%%%%%%%%%%%%%%%%%%%%%%%%%%%%%%%%
%%% ALGORITHM %%%
% \begin{minipage}{.7\linewidth}
% \centering
\begin{algorithm}[t]
\footnotesize
 \KwData{A binary file $\x$, target label $y$, conv size $c$}
 $k \gets c + (c - \textrm{len}(\x) ~ \textrm{mod} ~ c)$ ~~ \emph{// $k$ is the payload size}\\
 $\x^{\text{payload}} \sim \Un(0,N-1)^k$\\
 $\z^{\text{payload}}, ~ \z, ~\tilde{\z}^{\text{payload}} \leftarrow \M(\x^{\text{payload}}), ~\M(\x), ~\z^{\text{payload}}$\\
% $\x^{\text{new}}=\x \oplus \x^{\text{payload}}$\\
% $\z \leftarrow \M(\x)$\\
% \emph{// generate adversarial  in the embedding domain}  \\
% $\tilde{\z}^{\text{payload}} \leftarrow \z^{\text{payload}}$\\
 $\tilde{\z}^{\text{new}} \leftarrow [\z ; \tilde{\z}^{\text{payload}}]$\\ 
 \While{$\network_\theta(\tilde{\z}^{\text{new}}) > 0.5$}{
  $\tilde{\z}^{\text{payload}} \leftarrow \tilde{\z}^{\text{payload}} - \adveps \cdot \text{sign}\Big(\nabla_{\z} \sloss^*(\tilde{\z}^{\text{new}},\tilde{y} ;\theta)\Big)$\\
 $\tilde{\z}^{\text{new}} \leftarrow [\z ; \tilde{\z}^{\text{payload}}]$\\
 }
% \emph{// reconstruct the new binary file}  \\
 \For{$i\gets0$ \KwTo $\textrm{len}(\x)$}{
     $\tilde{\x}^{\text{new}}_i \leftarrow \argmin_j d(\tilde{\z}^{\text{new}}_i, M_j)$\\
 }
 return $\tilde{\x}^{\text{new}}$\\
 \caption{Adversarial Examples Generation }
 \label{algo:adv}
\end{algorithm}
% \end{minipage}
%%%%%%%%%%%%%%%%%%%%%%%%%%%%%%%%%%%%%%%%%%%%%%%%%%%%%%%%%%%%%%

\paragraph{Preserving Functionality}
\label{payload_teaser}
%So far we have discussed how to modify a binary file to cause misclassification.
%So far we have discussed how to modify the embedding vector and reconstruct a modified binary file.
Utilizing the above approach directly might not preserve the functionality of the file as the changes could affect the whole file.
To mitigate that we propose to contain modifications to a small chunk of bytes (payload).
%We propose to create a small chunk of bytes that are crafted so as to cause misclassification of the whole file. The modifications are contained to this small chunk only. 
%We refer to this chunk as the \emph{payload}.

This payload can be injected to the original file in one of two ways: 
(i) \emph{mid-file injection} -  the payload is placed in existing contingent unused bytes of sections where the physical size is greater than the virtual size;
(ii) \emph{end-of-file injection} - treating the payload as a new section and appending it to the file. Either approach would result in a valid and runnable code as the payload is inserted into non-executable code sections. We applied both methods and report results in \secref{sec:experiments}.

Algorithm~\ref{algo:adv} presents pseudo-code for generating a payload and injecting it at end-of-file. In this case, we append a uniformly random sequence of bytes of length $k$, $\smash{\x^{\text{payload}}\in\X^k}$, to the original file $\x$. We then embed the new binary $\smash{\x^{\text{new}}=[\x ; \x^{\text{payload}}]}$, to get $\smash{\z^{\text{new}}=[\z ; \z^{\text{payload}}]}$. Next, we iteratively perturb the appended segment $\smash{\z^{\text{payload}}}$ and stop when $\smash{f_\theta}$ misclassifies $\smash{\z^{\text{new}}}$. By perturbing only the appended segment, we ensure that $\z$ is kept intact and the functionality of $\x$ is preserved. This results in $\x^{\text{new}}$ that behaves identically to $\x$ but evades detection by $f_\theta$.
For the setting of mid-file injections we use the same perturbed payload $\smash{\x^{\text{payload}}}$ depicted in the previous paragraph, and inject it into contingent unused bytes where the physical size is greater than the virtual size.
% !TEX root = main.tex

\section{Experimental Results}
\label{sec:experiments}
\paragraph{Dataset}
We evaluated the effectiveness of our method using a dataset for the task of malware detection. Two sets of files were used:
(i) The first set is a collection of benign binary files that were collected as follows. We deployed a fresh Windows 8.1 32-bit image and installed software from over 50 different vendors using ``ninite'' \cite{ninite}.
We then collected all exe/dll with size 32KB or greater. This resulted in 7,150 files, which were labeled as \emph{benign};
(ii) The second set is a collection of 10,866 malicious binary files from 9 different malware families\footnote{Ramnit, Lollipop, Kelihos, Vundo, Simda, Tracur, Gatak, Obfuscator.ACY}. These files were taken from the \emph{Microsoft Kaggle 2015 dataset}\footnote{We thank Microsoft for allowing the use of their data for research.} \cite{kaggle} and were labeled as \emph{malicious}.
The entire data was shuffled and split into training, validation and test sets using 80/10/10 ratio respectfully.
Note that in the Kaggle 2015 dataset files were missing PE headers.
Therefore, for the generation of AEs, we used an unseen set of malicious binary files from \emph{VirusShare} with PE headers intact.

\paragraph{Detection Evasion}
\label{detection_evasion}
We trained a CNN E2E malware detector \cite{Raff2017} on the Kaggle 2015 dataset, until it reached a classification accuracy of 98\% on a validation set. This corresponded to accuracy of 97\% on the test set. The gradients of this model were later used for the the FGSM.
For the generation of AEs, we appended/injected to the file a payload denoted as $\x^{\text{payload}}$ that was initiated uniformly at random. Then, using iterative-FGSM, we perturbed the embedded representation of the payload, until the whole-file was misclassified. The FGSM procedure was implemented using $\smash{p=\infty}$ and $\smash{p=2}$. We evaluated the effectiveness of our method on the test set.
%We evaluated the evasion rate as a function of the loss tradeoff parameter $\alpha$ for each of the values of $p$.
%\subsection{Detection Evasion}
%\label{detection_evasion}
%As baseline for all experiments we use a value of $\smash{\alpha=0}$. This is the case where only a categorical-loss is used to generate AEs.
The injection procedure resulted in an evasion rate of 99.21\% and 98.83\% for $\smash{p=2}$ and $\smash{p=\infty}$ respectfully. These results were achieved under the constraint of retaining the original functionality and the payload length varied from 500 to 999 bytes (depending on the length of file modulo 500).
%Evasion rates of \FK{XXX}\% and \FK{XXX}\% were reached for $\smash{p=2}$ and $\smash{p=\infty}$ respectfully.
%We believe that the 12\% difference in evasion rate between $\smash{p=2}$ and $\smash{p=\infty}$ is due to poor reconstruction: the embeddings were perturbed to a point where they
%could not be associated with the fixed embeddings in $\M$. To further improve the evasion rate for $\smash{p=2}$, we plug our new loss function in the crafting of AEs.
%Recall that the loss tradeoff parameter $\alpha$ in \eqref{eq:3} weights the categorical loss and the embedding-similarity loss. 
%Where $\alpha=0$ means only the categorial loss active, whereas $\alpha=1$ means only the similarity-embedding loss is active. 
%We show the evasion rate as a function of $\alpha$ for $p=2$ and $p=\infty$ in \figref{fig:success_rate}. The highest evasion rate for $p=2$ was 95\% and was reached at $\smash{\alpha=0.5}$. The highest evasion rate for $p=\infty$ was 100\% and was reached for $\alpha \in [0,0.1]$.

 %% CONV ATTENTION %%%
 \begin{figure*}[h]
 \centering
 %\begin{subfigure}[t]{.3\textwidth}
 %  \centering
 %  \includegraphics[width=1\linewidth]{figures/success_rate}
 %  \caption{Evasion rate as function of the parameter $\alpha$ for $p$-norm of $p=2$ and $p=\infty$.}
 %  \label{fig:success_rate}
 %\end{subfigure}%
 ~~~
 \begin{subfigure}[t]{.4\textwidth}
   \centering
   \includegraphics[width=1\linewidth]{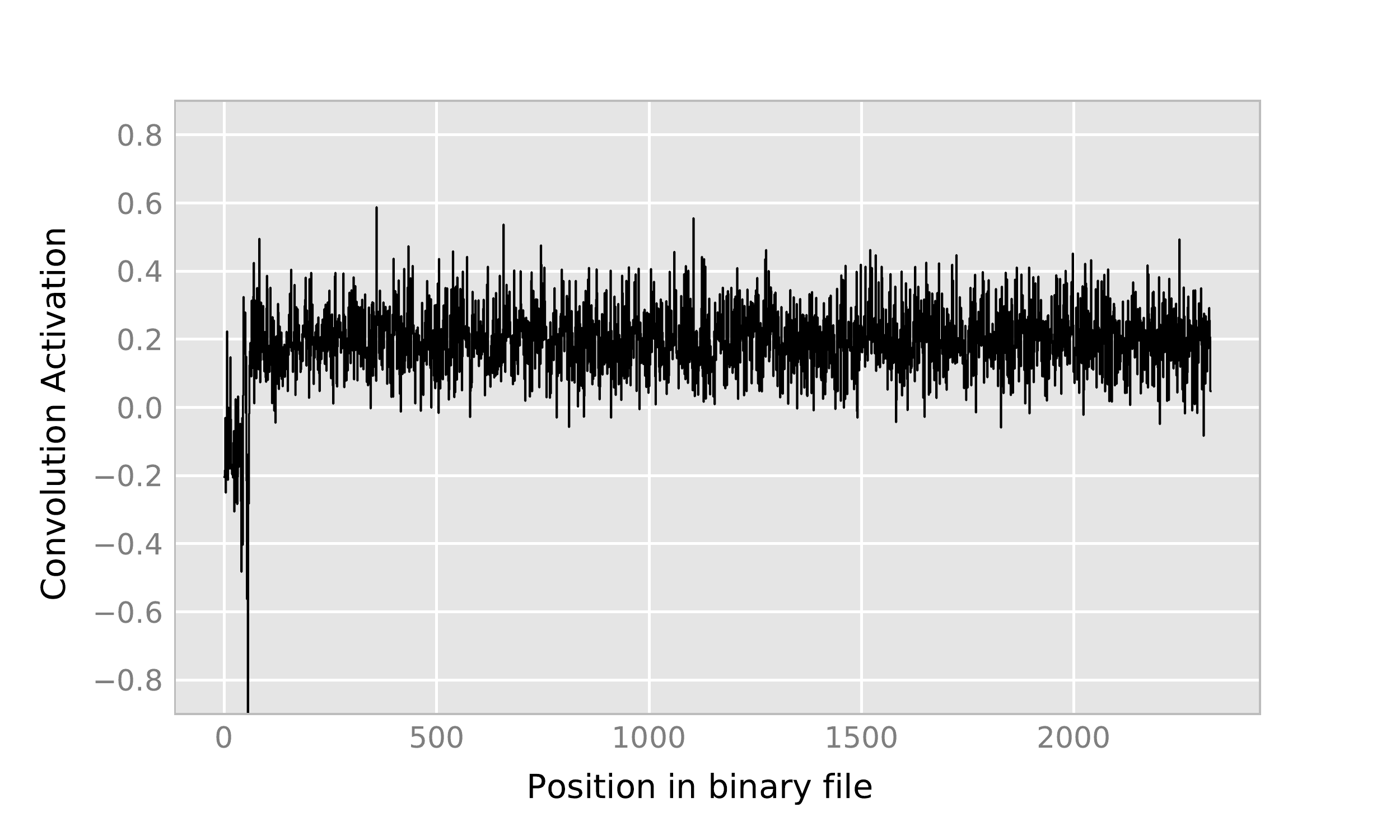}
   \caption{Average convolution activation.}
   \label{fig:conv_before}
 \end{subfigure}%
 ~~~
 \begin{subfigure}[t]{.4\textwidth}
   \centering
   \includegraphics[width=1\linewidth]{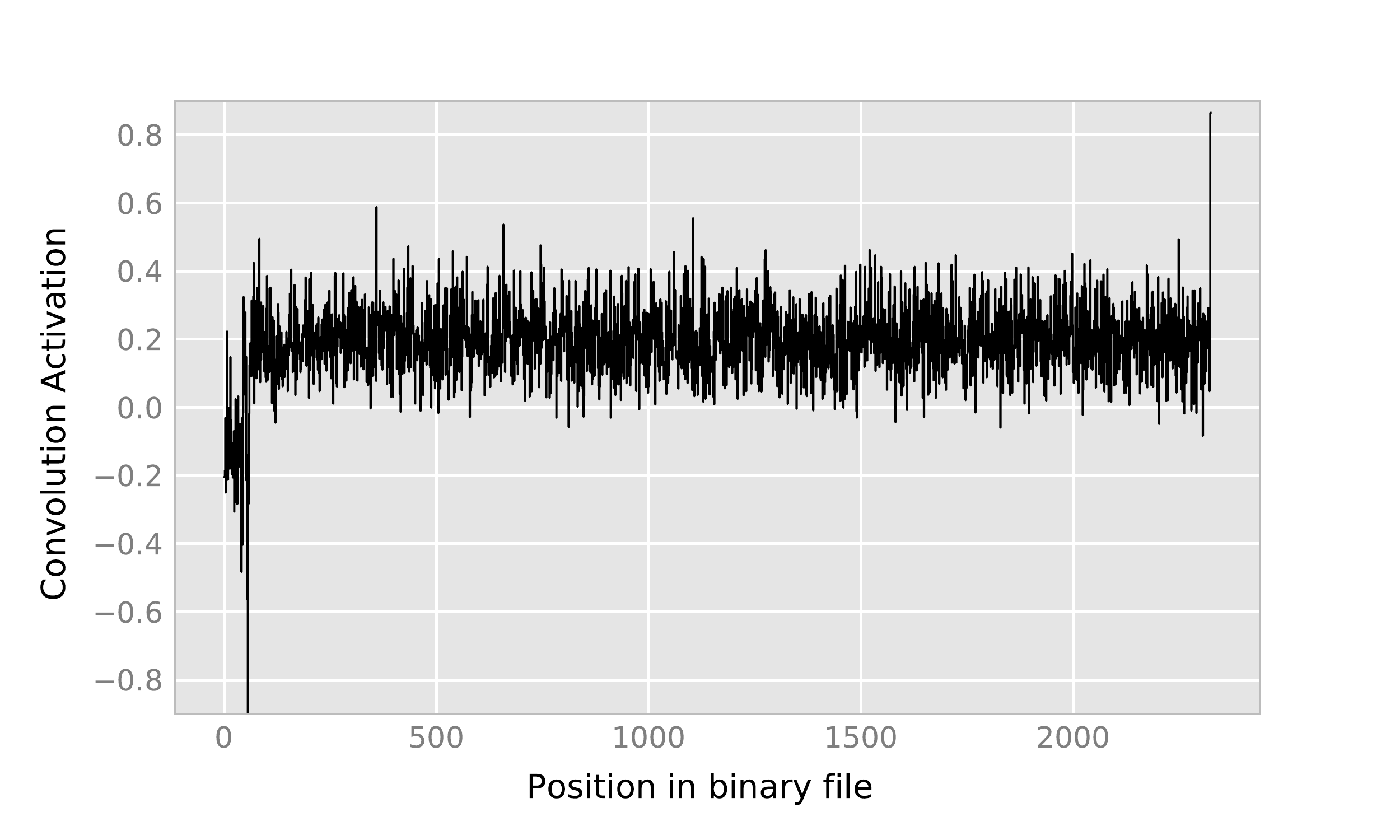}
   \caption{Average convolution activation.}
   \label{fig:conv_after}
 \end{subfigure}
 %\caption{``Attention shift'' demonstrated in convolutional activations~\cite{stock2017convnets}. The adversarial payload has higher activations on average than the rest of the file, essentially suppressing the rest of the malicious file due to temporal max-pooling.}
 \label{fig:general}
 \end{figure*}
 %%%%%%%%%%%%%%%%%%%%%%%%%%%%%%%%%%%%%%%%%%%%%%%%%%%%%%%%%%%%%

Evading detection can be explained by a shift in the model's attention. This phenomenon was previously demonstrated in \cite{stock2017convnets}.
To exemplify this point, we depicted the average activation of the convolutional layer along the position in a specific file before and after injecting the payload in \figref{fig:conv_after} and \figref{fig:conv_after} respectfully. Recall that in the \emph{MalConv} architecture, a temporal-max-pooling operation is performed after the convolution. It is seen from \figref{fig:conv_before} and \figref{fig:conv_after} that for the AE the max was achieved at the payload location, acting as a ``distraction'' to the malicious part of the code.

\paragraph{Payload}\label{payload_exp}
We now turn to introduce a set of experimental results to provide further analysis regarding the payloads' properties.
For these experiments we obtained 223 (previously-unseen) malicious binary files from 5 malware families: \emph{Gatak, Kelihos, Kryptic, Simda} and \emph{Obfuscator.ACY}. The original header was included to test the effectiveness of our method on files ``seen in the wild''.

\textit{Spatial invariance - }
\label{Spatial_Invariance}
We now demonstrate that the generated payload can fool the exploited network at multiple positions in the file. An attacker can benefit from such invariance as it allows for flexibility in the injection position.
For each file, we generated an adversarial version where the payload was appended at the end-of-file. Then, the injected payload was re-positioned to a mid-file position. 
%This position met two criteria: (i) it had sufficient unused contingent bytes where the physical size was greater than virtual size; and (ii) it was of multiple of $c \cdot a+\len(\x) \textrm{mod}~c$, where $a$ is an integer, $c$ is the convolution kernel size, and $\len(\x)$ is the size of the file.
To fit the payload inside the convolution window, the position was a multiple of $c a+\len(\x) \textrm{mod}~c$, where $a\in \N$, $c$ is the convolution kernel size, and $\len(\x)$ is the file size.
 We then evaluated the files using the exploited model before and after re-positioning. 100\% of files were still misclassified as benign. Meaning, the payload can be shifted to various positions and still fool the exploited model.

\textit{File transferability - }
We tested the transferability of the payload across different files: we injected a payload generated for file A to file B 
%(to positions meeting the criteria described in \ref{Spatial_Invariance})
. We then evaluated these files using the malware detector: 75\% were still misclassified as benign. Meaning, the same payload can be injected to multiple files and evade detection with high probability.

\textit{Payload size - }
Our payload size depends on $\smash{\len(x)\!\mod c}$, where $\len(x)$ is the size of the binary, and $c$ is the convolution kernel size. For the case of \emph{MalConv} our maximal payload size is 999 bytes. 
%Thus, the probability of finding files with sufficient space for the mid-file payload injection is higher for smaller payloads. 
To validate the minimalism property, we carried an attack using payload with maximal sizes of 1000, 1500, 2000, 2500 and recorded evasion rates of 99.02\%-99.21\% for $p=\infty$ and 98.04\%-98.83\% for $p=2$. We conclude that the evasion rate is not affected by the payload length given $c$.

\textit{Entropy - }
Entropy-based malware detectors flag suspicious sections by comparing their entropy with standard thresholds \cite{lyda2007using}. Regular text sections have an entropy of 4.5-5, entropy of above 6 is considered compressed/encrypted code.
%We measured the entropy of data sections for 9,500 benign files and 1000 malicious files. In benign files, \texttt{.data}, \texttt{.rdata}, \texttt{.idata} sections scored an entropy of 2.67, 4.57 and 4.39 respectfully. 
%In malicious files, \texttt{.data}, \texttt{.rdata}, \texttt{.idata} sections scored an entropy of 4.42, 5.15 and 4.7 respectfully. 
We analyzed the change in entropy caused by payload injection by calculating the entropy before and after injection. We found that the entropy change from 3.75 to 4.1 on average, and remained within an unsuspicious range.
%We analyzed the change in entropy of sections that were injected with a payload. For these sections, we calculated the entropy before and after the payload injection.
%We found that on average the entropy changed from 3.75 to 4.1 and remained within an unsuspicious range.
%We would like to stress that our goal is not to fool an entropy-based malware detector, but that our technique does not create new entropy suspicions. 
%We therefore conclude, that upon injection, our payload would not escalate the entropy of data sections.

\textit{Functionality Preserving - }\label{funtionality}
The functionality of the modified files is preserved by modifying non-executable sections. To further verify the functionality of the reconstructed binary files, we evaluated and compared the behavior graph of the files before and after the payload injection and found that they are identical.
%The functionality of the files is preserved by modifying non-executable sections. Nevertheless, one can claim that the functionality still needs to be validated.
%To that end, we have randomly sampled 10 files (5 benign, 5 malicious), each file was executed in a sandbox environment (\emph{Joe Sandbox}) and produced a ``behavior graph''. The behavior graph indicates actions taken by the executable. To validate the original functionality, we manually compared the behaviors graphs of the original files and the modified files. Results suggest that the behavior graphs are identical. Note that manual validation is extremely resource-intensive (as the behavior graphs are validated by a human), therefore, only 10 samples were evaluated.
% !TEX root = main.tex

\section{Future Work}
\label{sec:conclusion}

%%% QUICK RECAP %%%
%We have revisited the evaluation process of E2E deep learning for the task of malware detection. Our experiments show that AEs still pose a threat to such systems. 
%An analysis of the model showed an attention shift towards the appended/injected section payload. Essentially, the network ignored malicious code sections, and focused on the adversarial payload. 
%The generated payload was shown to posses properties that increase the detection evasion rate.
%Furthermore, we have proposed a new technique for generating AEs for the case of discrete sequences.

%%% FUTURE WORK %%%
In future work, we would like to decouple the generation process from the model's embedding layer to produce black-box attacks. We would also like to asses the reliability of real-world hybrid systems that use a wider array of detection techniques.

%%% END STATEMENT %%%
%As malware detection algorithms gradually take a more data-driven approach, bad-actors will seek new ways to avoid detection. Our work shows that a malicious binary file could be manipulated to avoid detection by such algorithms.
%While deep learning has shown significant breakthroughs in recent years, it seems critical to revisit the evaluation process of deep learning, especially in the case of security-sensitive domains in general, and for malware detection in particular.

\bibliographystyle{IEEEtran}
\bibliography{bibliography}

\end{document}